\documentclass[english]{article}
\usepackage[T1]{fontenc}
\usepackage[latin9]{inputenc}
\usepackage{geometry}
\usepackage{soul}
\usepackage{xcolor}
\usepackage{graphicx}
\usepackage{subfig}
\usepackage{authblk}
\makeatletter
\usepackage{url}

\makeatother

\usepackage{babel}
\begin{document}

\title{

%\includegraphics{research paper/images/tudelftlogo.png}~
%\\[5cm]
Using Autoencoders on Differentially Private Federated Learning GANs}

\author{Gregor Schram}
\author{Rui Wang}
\author{Kaitai Liang}
\affil{Delft University of Technology}

\maketitle
%\vfill
%\begin{center}
%A Dissertation Submitted to EEMCS faculty Delft University of Technology,\\
%In Partial Fulfilment of the Requirements\\
%For the Bachelor of Computer Science and Engineering
%\end{center}

%\newpage

\begin{abstract}
\noindent
Machine learning has been applied to almost all fields of computer science over the past decades. The introduction of GANs allowed for new possibilities in fields of medical research and text prediction. However, these new fields work with ever more privacy-sensitive data. In order to maintain user privacy, a combination of federated learning, differential privacy and GANs can be used to work with private data without giving away a users' privacy. Recently, two implementations of such combinations have been published: DP-Fed-Avg GAN and GS-WGAN. This paper compares their performance and introduces an alternative version of DP-Fed-Avg GAN that makes use of denoising techniques to combat the loss in accuracy that generally occurs when applying differential privacy and federated learning to GANs. We also compare the novel adaptation of denoised DP-Fed-Avg GAN to the state-of-the-art implementations in this field.
\end{abstract}

\section{Introduction}
Over the past two decades, machine learning and the broader field of artificial intelligence have grown exponentially. Nowadays, one may find these techniques applied everywhere, from fraudulent transaction detection by banks \cite{awoyemi2017credit} to self-driving cars and advanced autocorrect and typing aids on smartphones. The expanded usage of these systems however comes at a cost: data. In general, an increase in accuracy of a machine learning model requires an increase in training data. This is problematic for multiple reasons, one of the reasons being the absence of sufficiently large datasets and another being the privacy invasions when using certain datasets. 
%\newline

For this first problem, Ian Goodfellow \cite{goodfellow2014generative} and his colleagues came up with the technique of Generative Adversarial Nets (GANs) in 2014. This technique makes two machine learning models compete, with one being a discriminator and the other trying to generate examples of data that did not exist in the training set previously to fool the discriminator. This process in the end results in a much more accurate discriminator while requiring a much smaller dataset to begin with. 
%\newline

The second issue is one that has become more relevant in recent years as machine learning gets applied to more and more fields, the data it requires might be very privacy-sensitive. Think for example of medical data, pictures on a user's phone, or all the text input generated on someone's smartphone. Most people do not feel comfortable sharing this data with the large companies that often require it to improve their machine learning models. This is where Differential Privacy (DP) comes in, it was first brought up by Dwork and colleagues \cite{dwork2006calibrating} but some mathematics underlying the idea stem from Dalenius \cite{dalenius1977towards}. Differential privacy ensures that analysis of a dataset as a whole can happen while preserving the privacy of any individuals' data. 
%\newline

Another problem arising with the use of machine learning on private data, especially when this data is coming from phones, is that the dataset is not centralized. Conventional machine learning requires the dataset to be completely accessible at all times. This is not a good requirement when dealing with private and decentralized data, again like pictures or text input on a phone. This is where Federated Learning (FL) comes into the picture. Federated learning allows private and decentralized data to stay on edge devices, like a phone, while still benefiting from all the data available to improve the models globally. The term as used in this paper was originally put forward by researchers at Google \cite{mcmahan2017communication}. 

%\newline
Now that we understand the importance of differential privacy, federated learning and GANs we can also see why it is important to combine these. Differential privacy inherently makes a model less accurate which can be combatted by GANs, these however rely on centralized datasets which is why we need federated learning to apply it in real world use cases. The first paper combining these three techniques was only published two years ago by Sean Augenstein and his team at Google \cite{48690} with a real world use case being researched in the same year \cite{ramaswamy2020training}. 
%\newline

Since the convergence of these three techniques is very novel, this paper answers the question: Does adding a denoising step to DP-FL-GANs increase model accuracy while preserving privacy? This paper will first look at state-of-the-art techniques and add to them with a novel application of denoising for differential privacy. 
%\newline

This paper starts with an enumeration of related works in section 2. It then introduces a novel technique in section 3 and continues with an overview of the experiment setup in section 4. Section 5 gives an overview of the experiment results, while section 6 compares these to state-of-the-art implementations. Section 7 describes the ethics of this research, while section 8 contains the conclusions.

\section{Related Work}
While the field of differentially private Generative Adversarial Nets (GANs) is quite developed and well researched, the subfield of federated solutions is very new and not well researched yet. A few implementations do exist, and we will be comparing our solution to those. %\\
%\newline

\textbf{DP-Fed-Avg GAN \cite{48690}.} 
The first implementation and paper that proposed to apply differentially private GANs to a Federated Learning (FL) setting. The paper introduces the DP-Fed-Avg GAN algorithm, an adaptation of the Fed-Avg algorithm \cite{mcmahan2017learning}, to train both the discriminator and the generator in a federated setting while keeping user-level Differential Privacy (DP) guarantees under a trusted server. This trusted server is required because the averaging happens on the server and not on the edge devices. %\\
%\newline

\textbf{GS-WGAN \cite{chen2020gs}.}
This paper introduces a new way to use gradient information, enabling deeper models that can generate more useful samples representing a private dataset. Moreover, it allows this to be done both in a centralized and in a federated setting. In the federated setting, the proposed algorithm can provide user-level DP guarantees under an untrusted server, as gradient sanitization happens before sending the update. It also does not require rigorous hyperparameter tuning, something that DP-Fed-Avg GAN does require. 

\section{Proposed method}
This paper proposes to first apply Gaussian noise to local private training data and denoise it before training. Most Differential Privacy (DP) methods make use of Gaussian noise added to training data before training in order to enable user-level DP guarantees. The addition of noise inherently makes it more difficult for the Generative Adversarial Net (GAN) to generate and discriminate images reliably, the more noise one adds, the lower the accuracy of the GAN. Recently, there have been significant advances in noise removal on images \cite{zhang2017beyond} \cite{soh2021deep}. This noise removal will never lead back to the exact original image. This should preserve privacy as the original data is still permutated while resulting in higher accuracy due to more distinguishable training data. %\\
%\newline

\begin{figure}[h!]%
    \centering
    \subfloat[\centering Original sample]{{\includegraphics[width=3cm]{ 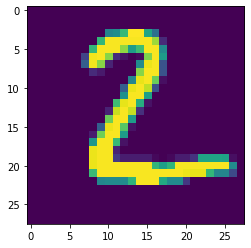} }}%
    \qquad
    \subfloat[\centering Sample with noise]{{\includegraphics[width=3cm]{ 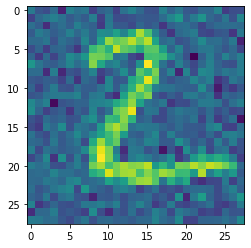} }}%
    \qquad
    \subfloat[\centering Sample after autodecoding]{{\includegraphics[width=3cm]{ 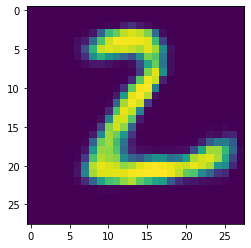} }}%
    \caption{An illustration of the effectiveness of an autoencoder}%
    \label{fig:autoencoder}%
\end{figure}

\label{denoising}The originally proposed denoising algorithm was to be based on the work by Majed El Helou and Sabine Susstrunk \cite{elhelou2020blind}. Later, we made the choice to focus on a simpler autoencoder, based on \cite{tensorflow}. We made this choice because our implementation works with the EMNIST dataset. This dataset is quite simple and uniform and thus does not benefit much from a very advanced algorithm as presented by \cite{elhelou2020blind}. The usage of an autoencoder also makes the experiment run quicker and more reliable. The effectiveness of an autoencoder can be seen in figure \ref{fig:autoencoder}. As we can see, most of the noise is removed, making it easier to recognize the shape of the sample and thus making the training of the GAN more effective.

%\noindent

\begin{figure}[h!]%
    \centering
    \includegraphics[width=4.8in]{ 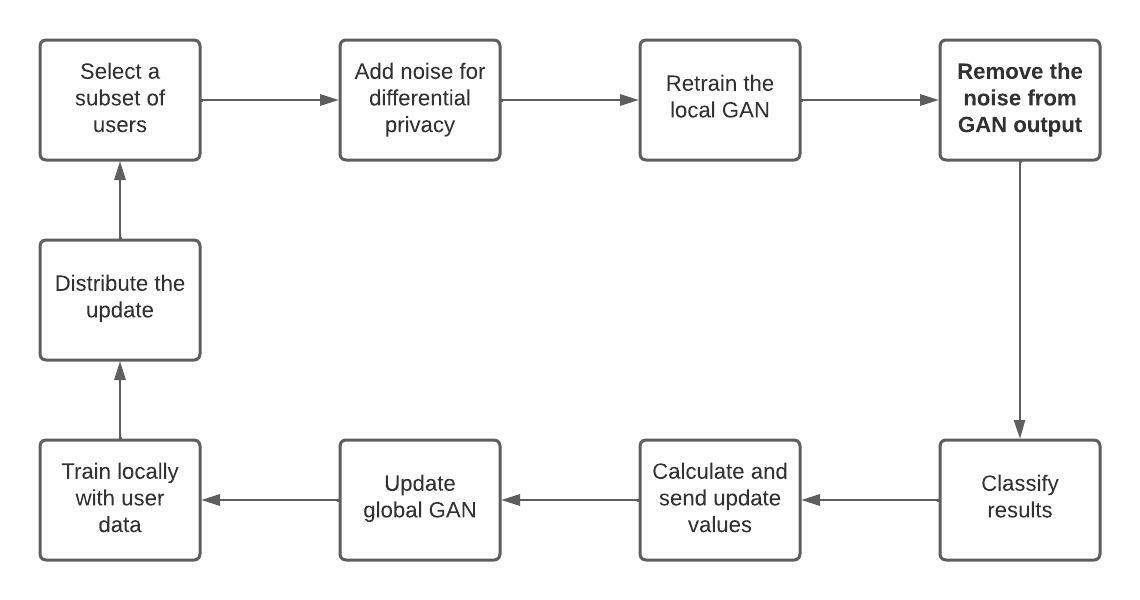}
    \caption{An overview of the steps in our modified DP-Fed-Avg GAN with noise removal}%
    \label{fig:flowchart}%
\end{figure}

As we can see in Figure \ref{fig:flowchart} our modified DP-Fed-Avg GAN algorithm goes through quite a few steps, the new step that we added has been highlighted. As one may understand from the flowchart, the removal of noise does not change the differential privacy level, as this happens after retraining the local GAN. Only when touching the input of the local GAN training session would this change.

\section{Experiment Setup}
For our experiment, we first looked at the currently existing methods of Differential Privacy (DP) Federated Learning (FL) Generative Adversarial Nets (GANs). As of writing, two exist: DP-Fed-Avg GAN \footnote{https://github.com/google-research/federated/tree/master/gans} and GS-WGAN \footnote{https://github.com/DingfanChen/GS-WGAN}. For both of these, the authors provided an implementation. These implementations were released with the intent to reproduce the results in the respective papers. Naturally, in order to compare these implementations against each other and against our implementation, we tried to reproduce the results produced in the respective papers \cite{48690} \cite{chen2020gs}. %\\
%\newline

We used \textit{conda} to create an environment for each of the solutions to run in isolation and to make sure that the setups were as close as possible to the ones used to generate the results in \cite{48690} and \cite{chen2020gs}. This ultimately allowed us to recreate some results of the DP-Fed-Avg GAN paper, but unfortunately we were not able to do this for the GS-WGAN paper as the provided code was incomplete and the federated setting was missing from it. %\\
%\newline

For the reproduction of the DP-Fed-Avg GAN results, we used the EMNIST dataset \cite{cohen2017emnist}, this is an extension of MNIST \cite{lecun-mnisthandwrittendigit-2010}, unlike MNIST it has letters as well as numbers and makes a distinction in capitalization, just like the original paper did. Since we are not interested in the practical use case of finding a bug, as introduced in \cite{48690} we did not introduce said bug in the experiments we ran. In order to do this, we selected only users with an accuracy over 93\%, and we set the image inversion probability to 0. Other than these values, we kept the exact same hyperparameters as used in the original paper. %\\
%\newline

For the new method proposed in section 3 we will make use of DP-Fed-Avg GAN. The reason we chose this algorithm as a base is that it is the only algorithm that has working code available, and implementing all the layers of GANs, FL and DP seemed unfeasible in the short timeframe of this research. Our method adds an extra step after each training round to remove noise on the intermediate results, which are also used in the next round, of the DP-Fed-Avg GAN. This extra step is based on the autoencoding algorithm described in \ref{denoising}. The code and setup instructions for our experiment can be found on GitHub\footnote{https://github.com/gregor160300/federated}. %\\
%\newline

As the original algorithm uses the EMNIST dataset, we also use this for our implementation. This requires some slight modification to the denoising algorithm, as it is designed for the MNIST dataset. We will use the same hyperparameters described earlier. However, this time we will use different noise levels as this affects the effectiveness of the denoising algorithm and the overall accuracy of our GAN.

\section{Results}
Our implementation has a very similar approach for applying differential privacy to a federated learning Generative Adversarial Net (GAN) as the DP-Fed-Avg GAN algorithm, we actually reuse most of the code from the DP-Fed-Avg GAN paper \cite{48690}. The only thing we added on top of the DP-Fed-Avg GAN algorithm is a step to denoise the intermediate outputs of the GAN. This did not touch the Differential Privacy (DP) level of the GAN, this was also not the intent. The intent was to reach a higher accuracy on classification of generated output, as well as reduce the Frechet Inception Distance (FID). Unfortunately, our experiment failed.
\begin{figure}[h!]%
    \centering
    \subfloat[\centering No DP and no noise removal]{{\includegraphics[width=3cm]{ 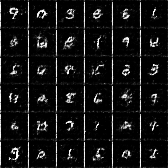} }}%
    \qquad
    \subfloat[\centering DP but no noise removal]{{\includegraphics[width=3cm]{ 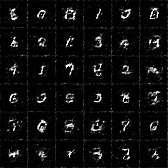} }}%
    \qquad
    \subfloat[\centering DP and noise removal]{{\includegraphics[width=3cm]{ 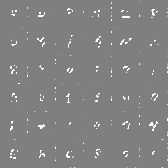} }}%
    \caption{An illustration of the impact of DP and denoising on the accuracy of the GAN}%
    \label{fig:ganoutputs}%
\end{figure}

Let us first take a look at the GAN output after 1000 rounds of training in figure \ref{fig:ganoutputs}. We see in figure \ref{fig:ganoutputs}a that the generated images when not using differential privacy nor noise removal are quite readable. In \ref{fig:ganoutputs}b we see a slight degradation in the readability due to the addition of DP guarantees. In the results of our implementation in \ref{fig:ganoutputs}c we see that all outputs have some sort of gray layer on top of them making it less readable and thus increasing the FID and lowering the accuracy of classification based on these generated results. %\\
%\newline

The way the autoencoder works for noise removal is to train a network on noisy input and non-noisy input. It then learns to remove noise from a noisy image by recognizing the differences. It is however trained on one specific noise level, in our case 20\%. This means that it will learn to reduce the noise best at 20\% noise, however our results might have more or less noise. The autoencoder will still try to remove the noise, but white noise on a black background then just becomes a gray filter on top of the entire image as it is unsure how to remove the noise. Our concept of noise removal might work with an algorithm that is agnostic of the noise level. \\

\section{Discussion}
The current implementations of DP-Fed-Avg GAN and GS-WGAN both have corresponding papers \cite{48690} \cite{chen2020gs} with results. We will focus on the model accuracy of the results and the differential privacy, as these are integral to our problem of increasing accuracy while preserving privacy. Differential privacy is measured by a value, \(\epsilon\) this value ideally is between 0 and 1, the lower, the better. The accuracy is measured by the Frechet Inception Distance (FID) as Generative Adversarial Nets (GANs) do not do classification, we check the similarity of the generated results to the training samples. The lower the distance between the two, the better. %\\
%\newline

We will first take a look at the claimed performance of the DP-Fed-Avg GAN. \cite{48690} claims that the DP-Fed-Avg GAN reaches an epsilon of \(9.99 \times 10^6\) in the simulation run in code. This is far above any useful privacy, as this basically means the chance that our GAN generates an actual training input (thus breaking privacy) is enormous. However, the researchers note that this is due to the small sample size of users (around 10 of a total of 3000) in the simulation. In a realistic scenario, with millions of users, they note an epsilon around the 1.4 mark. This is a much more practical privacy preservation level.
As for the FID, we see a claim around the 200 mark, which is not amazing, but considering the noise added for differential privacy, it is definitely acceptable.%\\
%\newline

The GS-WGAN paper \cite{chen2020gs} claims quite different results. The epsilon value they manage to reach is \(5.99 \times 10^2\), which is significantly lower than the value reached by the DP-Fed-Avg GAN. Note that this claim is also in the same simulated setup as the DP-Fed-Avg GAN, no claim is made however about the privacy level in a more realistic scenario. As for the FID, it is also significantly lower at just over 60, compared to the more than 200 with Fed-Avg GAN. Overall, GS-WGAN seems to be the more performant algorithm, both in terms of privacy preservation and GAN performance. %\\
%\newline

There is however one major sidenote to take on the GS-WGAN, that is that all these results should be taken at face value as there is no code available at the time of writing to reproduce these results. On the contrary, we were able to reproduce the results found in the DP-Fed-Avg GAN paper within a margin of error. The sidenote here being that we were only able to verify the performance of the simulation and not for the realistic setting. In summary, we thus have a quite performant unverified algorithm in GS-WGAN and a less performant, partially verified algorithm in DP-Fed-Avg GAN.
%\pagebreak

\begin{table}[hbt!]
\centering
\resizebox{\textwidth}{!}{%
\begin{tabular}{|l|l|l|l|}
\hline
& \textbf{GS-WGAN*} & \textbf{DP-Fed-Avg GAN} & \textbf{Ours} \\ \hline
\textbf{\begin{tabular}[c]{@{}l@{}}Frechet Inception Distance $\downarrow$ \end{tabular}} &
  $60$ &
  $0.2 * 10\textasciicircum{}3$ &
  $2.5 * 10\textasciicircum{}3$ \\ \hline
\textbf{\begin{tabular}[c]{@{}l@{}}Generator loss $\downarrow$ \end{tabular}}       & Unknown           & $-0.6$                    & $-0.5$          \\ \hline
\textbf{\begin{tabular}[c]{@{}l@{}}Classifier accuracy $\uparrow$ \end{tabular}} & Unknown           & $60\%$                    & $18\%$          \\ \hline
\textbf{\begin{tabular}[c]{@{}l@{}}Epsilon $\downarrow$ \end{tabular}} &
  $5.99 * 10\textasciicircum{}2$ &
  $9.99 * 10\textasciicircum{}6**$ &
  $9.99 * 10\textasciicircum{}6**$ \\ \hline
\end{tabular}
}
\caption{A performance comparison of our algorithm against GS-WGAN and DP-Fed-Avg GAN. \textit{* claimed, but unverified values used, ** experimental setting value, lower in real world scenarios}}
\label{tab:comparison}
\end{table}

When we compare the performance of our method of applying an autoencoder to the DP-Fed-Avg GAN, we unfortunately see worse performance than both DP-Fed-Avg GAN and GS-WGAN. As we can see in Table \ref{tab:comparison} the Epsilon value, which indicates the differential privacy level, stays the same from DP-Fed-Avg GAN to our modification of it, however the FID as well as the classifier accuracy is significantly lower. These lower values are most likely caused by the issue described in the Results section. %\\

\begin{figure}[h!]
    \centering
    \includegraphics[width=6cm]{ 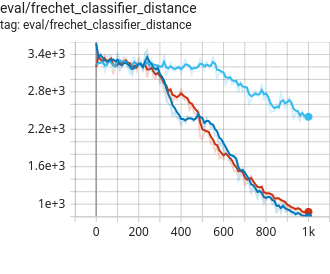}
    \caption{Frechet Inception Distance: Red = Fed-Avg GAN without DP, Light Blue = Ours, Blue = Fed-Avg GAN with DP}
    \label{fig:fid}%
\end{figure}

%\noindent
If we look into more detail at the performance over the 1000 iterations, we can see in Figure \ref{fig:fid} that the further we proceed, the lower our FID becomes. This makes sense intuitively, as the GAN will get better at creating results similar to its input with more training. What we also see is that this improvement goes much slower with our implementation than the others. This is logical, considering a gray blur is added in our results due to the failing denoising algorithm. In future research, one might try to run the algorithm for more iterations to see if the FID slowly keeps declining or plateaus. %\\

\begin{figure}[h!]%
    \centering
    \subfloat[\centering Generator Loss: Red = Fed-Avg GAN without DP, Light Blue = Ours, Blue = Fed-Avg GAN with DP]{{\includegraphics[width=6cm]{ 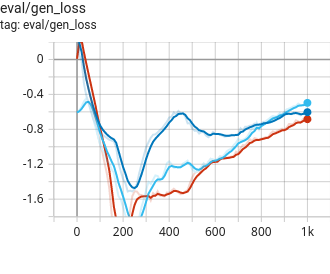} }}%
    \qquad
    \subfloat[\centering Classifier score: Red = Fed-Avg GAN without DP, Light Blue = Ours, Blue = Fed-Avg GAN with DP]{{\includegraphics[width=6cm]{ 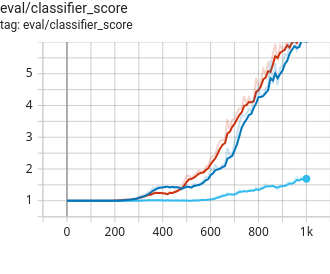} }}%
    \caption{Generator loss and classifier score}
    \label{fig:loss_and_classifier}%
\end{figure}

What is interesting to see is that the generator loss (see Figure \ref{fig:loss_and_classifier}a for our algorithm does not seem to be markedly higher or lower than the regular DP-Fed-Avg GAN. Unfortunately, we could not find a clear reason for this behavior. When we look at the classifier score (see Figure \ref{fig:loss_and_classifier}b), remember a GAN has usually trains against a classifier, we see a notable decrease in accuracy with our implementation due to previous described issues with our experiment. Here again, it might be interesting to see how the accuracy progresses after 1000 rounds when doing future research into optimization of the DP-Fed-Avg GAN algorithm.

\section{Conclusion}
This paper introduced the idea of using autoencoders on differentially private federated learning GANs. This novel approach applied an autoencoder on top of the DP-Fed-Avg GAN. The idea is to reduce the noise on the GAN output and thus improve the readability of the output, making it easier to classify or spot weird outputs.  Unfortunately, due to the autoencoder being trained at a specific noise level and the differential privacy calculations adding arbitrary levels of noise, this technique failed. The concept could still work when tried with a noise removal algorithm that does not require knowing the noise level upfront. 
%\newline

This paper also compared the performance of the novel approach to the existing GS-WGAN and DP-Fed-Avg GAN algorithms. These results showed that both in terms of privacy and generated results, the GS-WGAN algorithm is currently the best. The paper also addressed some questions with regard to the ethics of this research and the societal impact thereof.

\bibliographystyle{ieeetr}
\bibliography{main}

\end{document}